\pdfoutput=1
\documentclass[11pt,a4paper]{article}
\usepackage[hyperref]{naaclhlt2018}
\usepackage{times}
\usepackage{latexsym}
\usepackage{url}
\usepackage{footmisc}

\aclfinalcopy

\title{IIIDYT at SemEval-2018 Task 3: Irony detection in English tweets}

\author{Edison Marrese-Taylor\textsuperscript{1}\textsuperscript{*}, Suzana Ilic\textsuperscript{2}\textsuperscript{*}, Jorge A. Balazs\textsuperscript{1}\textsuperscript{*}, \\ {\bf Helmut Prendinger\textsuperscript{2}, Yutaka Matsuo\textsuperscript{1}}\\
Graduate School of Engineering, The University of Tokyo, Japan\textsuperscript{1}\\
emarrese,jorge,matsuo@weblab.t.u-tokyo.ac.jp\\
National Institute of Informatics, Tokyo, Japan\textsuperscript{2}\\
suzana,helmut@nii.ac.jp\\
\footnotesize{\textsuperscript{*}Authors contributed equally to this work.}}

\date{}

\begin{document}

\maketitle

\begin{abstract} 
In this paper we introduce our system for the task of Irony detection in English tweets, a part of SemEval 2018. We propose representation learning approach that relies on a multi-layered bidirectional LSTM, without using external features that provide additional semantic information. Although our model is able to outperform the baseline in the validation set, our results show limited generalization power over the test set. Given the limited size of the dataset, we think the usage of more pre-training schemes would greatly improve the obtained results.
\end{abstract}

\section{Introduction}

Sentiment analysis and emotion recognition, as two closely related subfields of affective computing, play a key role in the advancement of artificial intelligence \cite{Cambria2017}. However, the complexity and ambiguity of natural language constitutes a wide range of challenges for computational systems.

In the past years irony and sarcasm detection have received great traction within the machine learning and NLP community \cite{Joshi2016}, mainly due to the high frequency of sarcastic and ironic expressions in social media. Their linguistic collocation inclines to flip polarity in the context of sentiment analysis, which makes machine-based irony detection critical for sentiment analysis \cite{Poria2016,VanHee2015}. Irony is a profoundly pragmatic and versatile linguistic phenomenon. As its foundations usually lay beyond explicit linguistic patterns in re-constructing contextual dependencies and latent meaning, such as shared knowledge or common knowledge \citep{Joshi2016}, automatically detecting it remains a challenging task in natural language processing. 

In this paper, we introduce our system for the shared task of Irony detection in English tweets, a part of the 2018 SemEval \cite{VanHee2016SemEval}. We note that computational approaches to automatically detecting irony often deploy expensive feature-engineered systems which rely on a rich body of linguistic and contextual cues \citep{Bamman2015a,Joshi2015}. The advent of Deep Learning applied to NLP has introduced models that have succeeded in large part because they learn and use their own continuous numeric representations \cite{hinton1984distributed} of words \cite{mikolov_distributed_2013}, offering us the dream of forgetting manually-designed features. To this extent, in this paper we propose a representation learning approach for irony detection, which relies on a bidirectional LSTM and pre-trained word embeddings.


\section{Data and pre-processing}

For the shared task, a balanced dataset of 2,396 ironic and 2,396 non-ironic tweets is provided. The ironic corpus was constructed by collecting self-annotated tweets with the hashtags \textit{\#irony}, \textit{\#sarcasm} and \textit{\#not}. The tweets were then cleaned and manually checked and labeled, using a fine-grained annotation scheme \citep{VanHee2015}. The corpus comprises different types of irony:

\begin{itemize}
\item Verbal irony (polarity contrast): 1,728 instances 
\item Other types of verbal irony: 267 instances.
\item Situational irony: 401 instances
\end{itemize}

Verbal irony is often referred to as an utterance that conveys the opposite meaning of what of literally expressed \citep{Grice1975, Wallace2015}, e.g. \emph{I love annoying people}. Situational irony appears in settings, that diverge from the expected \citep{Lucariello1994}, e.g. \emph{an old man who won the lottery and died the next day}. The latter does not necessarily exhibit polarity contrast or other typical linguistic features, which makes it particularly difficult to classify correctly.

For the pre-processing we used the Natural Language Toolkit \cite{Loper2002}. As a first step, we removed the following words and hashtagged words: \emph{not}, \emph{sarc}, \emph{sarcasm}, \emph{irony}, \emph{ironic}, \emph{sarcastic} and \emph{sarcast}, in order to ascertain a clean corpus without topic-related triggers. To ease the tokenizing process with the NLTK TweetTokenizer, we replaced two spaces with one space and removed usernames and urls, as they do not generally provide any useful information for detecting irony.

We do not stem or lowercase the tokens, since some patterns within that scope might serve as an indicator for ironic tweets, for instance a word or a sequence of words, in which all letters are capitalized \cite{Tsur2010}.

\section{Proposed Approach}

The goal of the subtask A was to build a binary classification system that predicts if a tweet is ironic or non-ironic. In the following sections, we first describe the dataset provided for the task and our pre-processing pipeline. Later, we lay out the proposed model architecture, our experiments and results.

\subsection{Word representation}

Representation learning approaches usually require extensive amounts of data to derive proper results. Moreover, previous studies have shown that initializing representations using random values generally causes the performance to drop. For these reasons, we rely on pre-trained word embeddings as a means of providing the model the adequate setting. We experiment with GloVe\footnote{\url{nlp.stanford.edu/projects/glove}} \cite{Pennington2014} for small sizes, namely 25, 50 and 100. This is based on previous work showing that representation learning models based on convolutional neural networks perform well compared to traditional machine learning methods with a significantly smaller feature vector size, while at the same time preventing over-fitting and accelerates computation (e.g \citep{Poria2016}. 

GloVe embeddings are trained on a dataset of 2B tweets, with a total vocabulary of 1.2 M tokens. However, we observed a significant overlap with the vocabulary extracted from the shared task dataset. To deal with out-of-vocabulary terms that have a frequency above a given threshold, we create a new vector which is initialized based on the space described by the infrequent words in GloVe. Concretely, we uniformly sample a vector from a sphere centered in the centroid of the 10\% less frequent words in the GloVe vocabulary, whose radius is the mean distance between the centroid and all the words in the low frequency set. For the other case, we use the special \textit{UNK} token.

To maximize the knowledge that may be recovered from the pre-trained embeddings, specially for out-of-vocabulary terms, we add several token-level and sentence-level binary features derived from simple linguistic patterns, which are concatenated to the corresponding vectors.

\paragraph{Word-level features}
\begin{enumerate}
	\item If the token is fully lowercased.
    \item If the Token is fully uppercased.
    \item If only the first letter is capitalized.
    \item If the token contains digits.
\end{enumerate}

\paragraph{Sentence-level features}
\begin{enumerate}
	\item If any token is fully lowercased.
    \item If any token is fully uppercased. 
    \item If any token appears more than once.
\end{enumerate} 


\subsection{Model architecture}

Recurrent neural networks are powerful sequence learning models that have achieved excellent results for a variety of difficult NLP tasks \citep{IanGoodfellowYoshuaBengio2017}. In particular, we use the last hidden state of a bidirectional LSTM architecture \cite{Hochreiter1997} to obtain our tweet representations. This setting is currently regarded as the state-of-the-art \cite{Barnes2017} for the task on other datasets. To avoid over-fitting we use Dropout \cite{srivastava2014dropout} and for training we set binary cross-entropy as a loss function. For evaluation we use our own wrappers of the the official evaluation scripts provided for the shared tasks, which are based on accuracy, precision, recall and F1-score.

\section{Experimental setup}

Our model is implemented in PyTorch \cite{Paszke2017}, which allowed us to easily deal with the variable tweet length due to the dynamic  nature of the platform. We experimented with different values for the LSTM hidden state size, as well as for the dropout probability, obtaining best results for a dropout probability of $0.1$ and $150$ units for the the hidden vector. We trained our models using 80\% of the provided data, while the remaining 20\% was used for model development. We used Adam \cite{kingma_adam2015}, with a learning rate of $0.0001$ and early stopping when performance did not improve on the development set. Using embeddings of size 100 provided better results in practice. Our final best model is an ensemble of four models with the same architecture but different random initialization.

To compare our results, we use the provided baseline, which is a non-parameter optimized linear-kernel SVM that uses TF-IDF bag-of-word vectors as inputs. For pre-processing, in this case we do not preserve casing and delete English stopwords.

\section{Results}
To understand how our strategies to recover more information from the pre-trained word embeddings affected the results, we ran ablation studies to compare how the token-level and sentence-level features contributed to the performance. Table \ref{table:ablation_results_summary} summarizes the impact of these features in terms of F1-score on the validation set. 

\begin{table}[!h]
	\centering
    \begin{tabular}{c | c | c}
    \textbf{Feature} &  \textbf{Yes} & 	\textbf{No} \\
    \hline
    Token-level    & 0.6843 & 0.7008 \\
    Sentence-level & 0.6848 & 0.6820
    \end{tabular}
\caption{Results of our ablation study for binary features in terms of F1-Score on the validation set.}
\label{table:ablation_results_summary}
\end{table}

We see that sentence-level features had a positive yet small impact, while token-level features seemed to actually hurt the performance. We think that since the task is performed at the sentence-level, probably features that capture linguistic phenomena at the same level provide useful information to the model, while the contributions of other finer granularity features seem to be too specific for the model to leverage on.

Table \ref{table:results_summary} summarizes our best single-model results on the validation set (20\% of the provided data) compared to the baseline, as well as the official results of our model ensemble on the test data.

\begin{table}[!h]
	\centering
    \scriptsize
    \begin{tabular}{c | c | c | c | c}
    \textbf{Split} &  \textbf{Accuracy} &	\textbf{Precision} & \textbf{Recall} & 	\textbf{F1-score} \\
    \hline
    Baseline Valid  & 0.6375 	 &  0.6440 &  	0.6096 &    0.6263 \\
    Ours Valid 		& 0.6610	 &	0.6369 &	0.8447 &	0.7262 \\
    Ours Test  		& 0.3520	 &	0.2568 &	0.3344 &	0.2905
    \end{tabular}
\caption{Summary of the obtained best results on the valid/test sets.}
\label{table:results_summary}
\end{table}

Out of 43 teams our system ranked 421st with an official F1-score of 0.2905 on the test set. Although our model outperforms the baseline in the validation set in terms of F1-score, we observe important drops for all metrics compared to the test set, showing that the architecture seems to be unable to generalize well. We think these results highlight the necessity of an ad-hoc architecture for the task as well as the relevance of additional information. The work of \citet{Felbo2017a} offers interesting contributions in these two aspects, achieving good results for a range of tasks that include sarcasm detection, using an additional attention layer over a BiLSTM like ours, while also pre-training their model on an emoji-based dataset of 1246 million tweets. 

Moreover, we think that due to the complexity of the problem and the size of the training data in the context of deep learning better results could be obtained with additional resources for pre-training. Concretely, we see transfer learning as one option to add knowledge from a larger, related dataset could significantly improve the results \citep{Pan2010}. Manually labeling and checking data is a vastly time-consuming effort. Even if noisy, collecting a considerably larger self-annotated dataset such as in \citet{Khodak2017} could potentially boost model performance. 

\section{Conclusion}

In this paper we presented our system to SemEval-2018 shared task on irony detection in English tweets (subtask A), which leverages on a BiLSTM and pre-trained word embeddings for representation learning, without using human-engineered features. Our results showed that although the generalization capabilities of the model are limited, there are clear future directions to improve. In particular, access to more training data and the deployment of methods like transfer learning seem to be promising directions for future research in representation learning-based sarcasm detection.

\renewcommand*{\bibfont}{\footnotesize}
\bibliography{Sarcasm}
\bibliographystyle{acl_natbib}
\end{document}